\documentclass[sigconf,nonacm]{acmart}

\usepackage{amsmath}
\usepackage{booktabs}
\usepackage{tabularx}
\usepackage{array}

\AtBeginDocument{%
  }

\setcopyright{none}
\renewcommand\footnotetextcopyrightpermission[1]{}

\newcolumntype{Y}{>{\raggedright\arraybackslash}X}
\newcolumntype{Z}{>{\centering\arraybackslash}X}

\begin{document}

\title[Dear Algo: A Precision-First Agentic Intent Layer]%
  {Dear Algo: A Precision-First Agentic Intent Layer for Unified Search and Recommendation}

\author{Rui Wang, Jiazhou Wang, Zheng Wei, Chenglin Lu, Fangcheng Sun, Ivy Sun, Jin Sun, Hui Geng, Lillian Zhang, Chao Yang, Lei Chen, Shahin Sefati, Reem Helou, Joe Zhou, Babak Shakibi, Yiyi Pan, Bi Xue, Hong Yan\textsuperscript{*}, Shujian Bu}
\affiliation{%
  \institution{Meta Platforms Inc.}
  \country{USA}
}

\renewcommand{\shortauthors}{Rui Wang et al.}

\begin{abstract}
Search and recommendation serve a shared discovery objective but encode intent
differently. We study this boundary through Dear Algo on Threads, a deployed
product where open-ended requests such as \emph{more NBA news} or \emph{less
politics} steer subsequent feed recommendations rather than return a one-shot
result list. Its agentic intent layer compiles explicit, inferred, negative, and
compound intent into a grounded executable plan, then invokes conventional
retrieval and optional semantic or multimodal reranking. The layer shares an
intent-to-retrieval contract without requiring one model or serving path across
search-like and recommendation-like modes.

We evaluate Dear Algo under a precision-first objective. In a blinded audit of
300 public request-item pairs (296 evaluable), a strict categorical
LLM-as-a-judge gate achieved 94.4\% exact-Relevant precision [88.8\%, 98.9\%].
Across 72 normalized request clusters, the full configuration produced 7.73
judge-qualified candidates per 20 slots versus 6.61 for an LLM-derived-query
baseline, a gain of 1.11 [0.12, 2.12]. In a candidate-randomized serving-path
study restricted to the reranker path's first 72 eligible hours, the user-weighted
judge-Irrelevant share among judged admissions was 2.80\% versus 4.78\% off
(-1.97 points [-3.02, -0.94]), while Exact-Relevant share was 2.24 points
higher [0.08, 4.41].

Together, these studies show how explicit natural-language intent can be
carried into feed recommendation under a precision-first evaluation framework.
\end{abstract}

\ccsdesc[500]{Information systems~Recommender systems}
\ccsdesc[500]{Information systems~Retrieval models and ranking}
\ccsdesc[300]{Information systems~Evaluation of retrieval results}
\ccsdesc[300]{Computing methodologies~Natural language processing}

\keywords{unified search and recommendation, agentic retrieval, intent
understanding, retrieval-augmented generation, structured retrieval,
precision-first evaluation, industrial recommender systems}

\maketitle
\begingroup
\renewcommand{\thefootnote}{\fnsymbol{footnote}}
\footnotetext[1]{Also with Google DeepMind, USA.}
\endgroup

\section{Introduction}

Search exposes explicit but typically short-lived intent through a query and
is commonly evaluated by query-item relevance. Recommendation builds
longer-lived preferences primarily from implicit actions such as views,
shares, skips, and replies and is often evaluated through engagement. Even
over a shared catalog, an explicit request is rarely converted into durable,
executable state for feed recommendation. LLMs enable open-ended
interpretation, but production retrieval still requires valid identifiers,
explicit constraints, bounded execution, fallback, and inspectable telemetry.

\begin{figure*}[t]
  \centering
  \begin{minipage}[t]{0.49\textwidth}
    \centering
    \includegraphics[width=\linewidth]{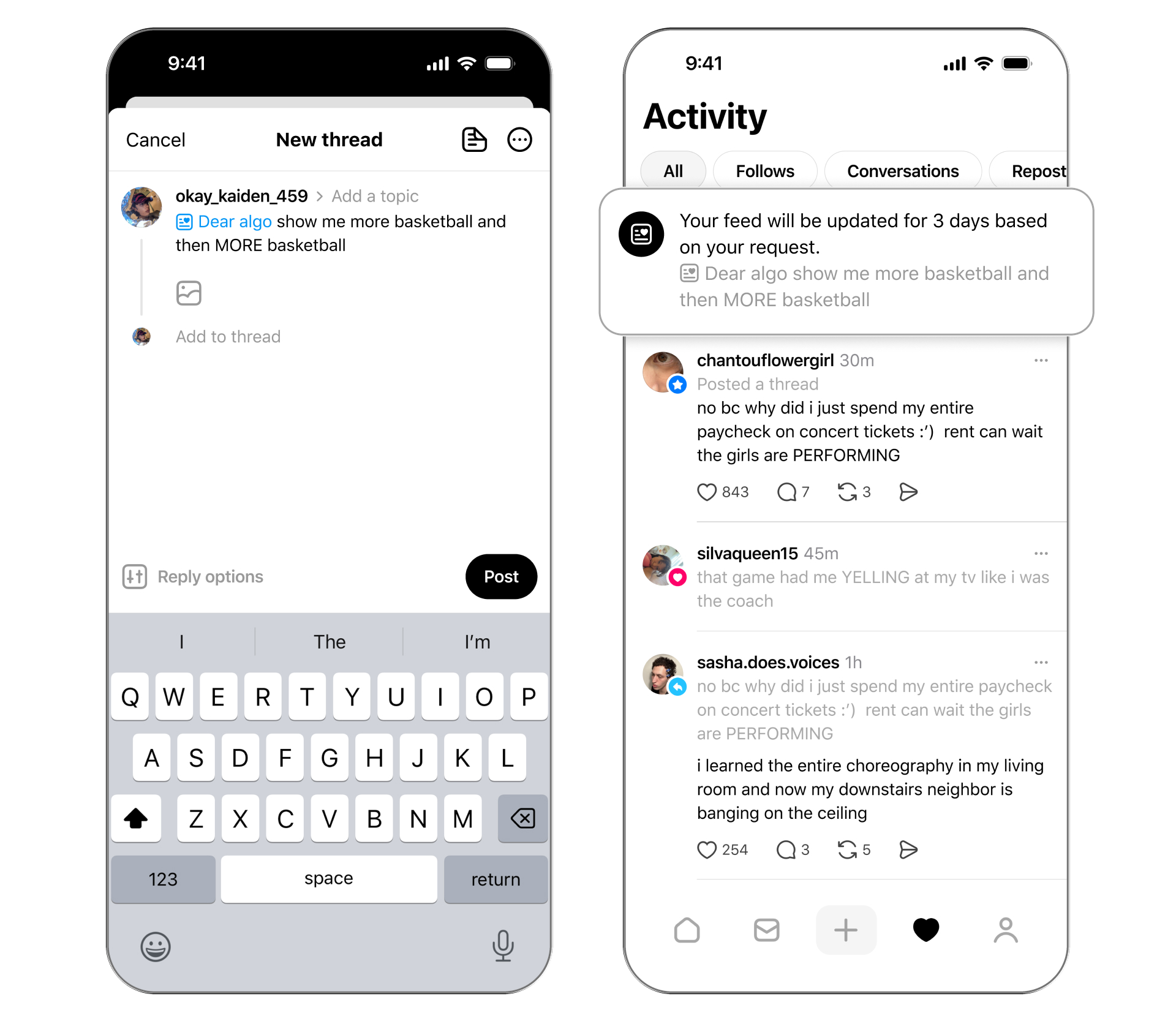}
    \small (a) Natural-language feed steering
  \end{minipage}\hfill
  \begin{minipage}[t]{0.49\textwidth}
    \centering
    \includegraphics[width=\linewidth]{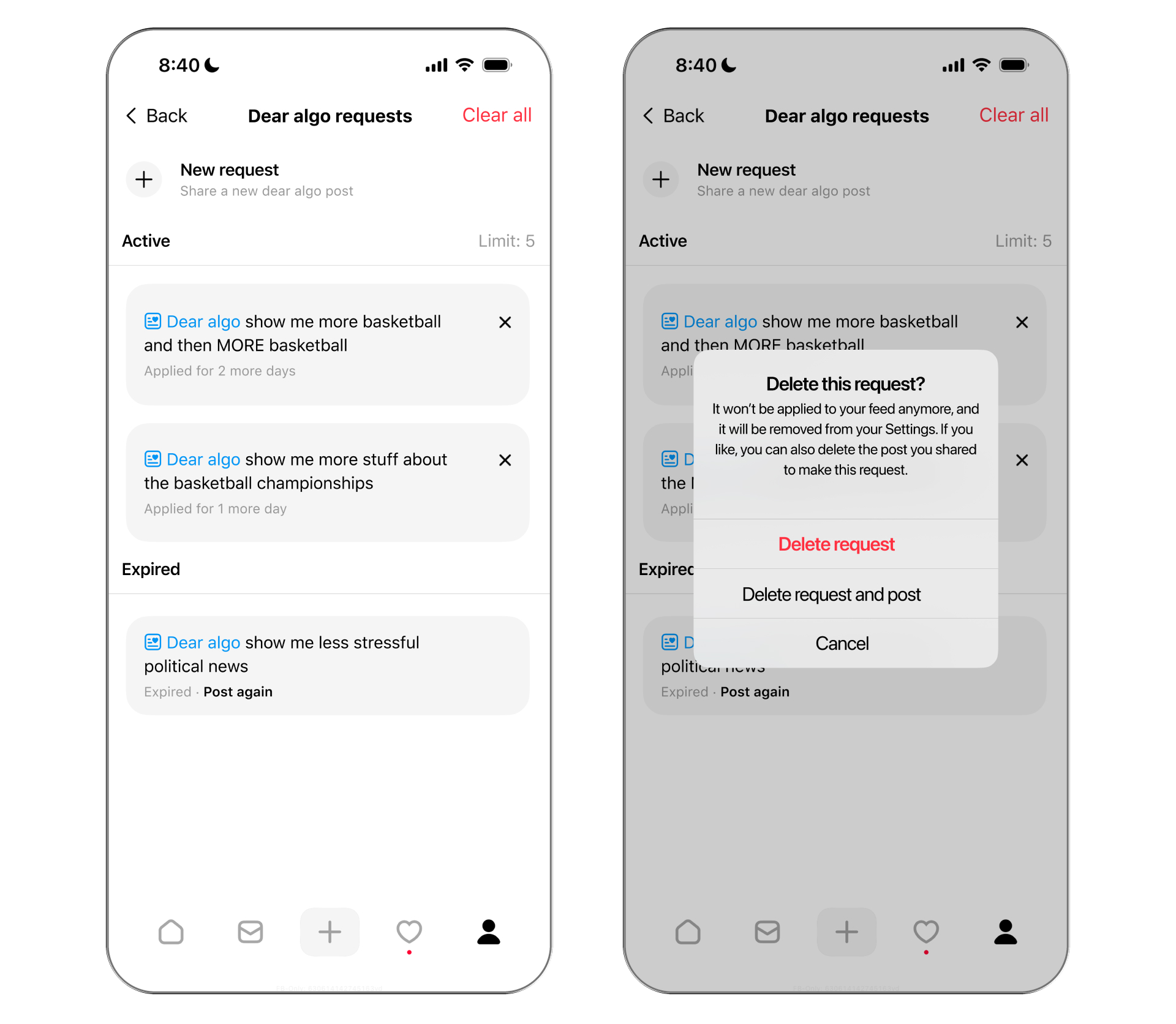}
    \small (b) Persistent preference settings
  \end{minipage}
  \caption{Dear Algo product surfaces on Threads. Users express feed
  preferences through natural-language requests and review or manage the
  resulting controls in Settings. Images reproduced from the public Meta
  Newsroom announcement~\cite{meta-dear-algo}.}
  \Description{Two public Threads product images. The left panel shows the
  Dear Algo request surface, and the right panel shows the related settings
  surface for persistent feed preferences.}
  \label{fig:dear-algo-product}
\end{figure*}

Dear Algo is a deployed Threads interface at this boundary. A user can ask for
\emph{more NBA news}, \emph{less politics}, or content that will
\emph{make me laugh}; the instruction steers subsequent feed retrieval rather
than returning only a one-shot ranked list~\cite{meta-dear-algo,
verge-dear-algo,techcrunch-threads-personalization}.
Figure~\ref{fig:dear-algo-product} shows the public request and settings
surfaces.

Dear Algo uses an agentic intent layer built on the SilverTorch serving
substrate~\cite{silvertorch}. The layer interprets an instruction into typed
positive and negative constraints, grounds open-ended phrases to a controlled
item-side semantic vocabulary, and compiles a reusable
SilverTorch Query Language (STQL) plan for candidate retrieval and Boolean
filtering. STQL is an internal JSON-serialized retrieval DSL, not relational
SQL. Optional semantic or multimodal reranking operates on a bounded candidate
set.

The compiler and plan are shared, while interfaces, operating points, and
serving paths may remain mode-specific. This modularity matters because the
product computational cost is asymmetric: an irrelevant item admitted into a persistent
feedback loop can cause repeated unwanted exposure, whereas rejection mainly
reduces coverage. We therefore separate admission precision, recall, ranking and engagement, and then report user actions accordingly.

This paper contributes:
\begin{enumerate}
  \item \textbf{A shared executable intent layer for feed steering.} Dear
    Algo maps explicit requests and inferred preferences into one grounded plan
    while retaining mode-specific serving paths.
  \item \textbf{A layered account of unification.} We distinguish
    infrastructure, representation, and empirical unification and identify
    which layers the system is designed and implemented to share.
  \item \textbf{A precision-first evaluation bridge.} We combine blinded human
    calibration, offline candidate evaluation, randomized admitted-set
    analysis, and behavioral analysis
    without treating their estimands as interchangeable.
  \item \textbf{Industrial evidence and measurement lessons.} We report human
    calibration, paired offline retrieval, candidate-randomized admitted-set
    quality, and observational usage, and identify the logging required for
    causal cross-mode evaluation.
\end{enumerate}

\section{Related Work}

\subsection{Unified Search and Recommendation}

Search and recommendation address a common discovery problem over shared
catalogs, but condition their decisions on different evidence: search starts
from an explicit, often transient query, whereas recommendation infers
preferences from interaction history. Existing work unifies behavior
sequences, graphs, encoders, or transition models
\cite{user,zhao2022srjgraph,unifiedssr,unisar}, and KuaiSAR provides aligned
logs for studying both tasks \cite{kuaisar}. Other approaches share item
representations \cite{zamani2018joint,zamani2020joint}, transfer search
representations into recommendation \cite{searchmeetsrec,zhang2024uditsr}, or
use generative approaches to jointly model both tasks
\cite{bridging,shi2025gensar,liao2026minsar}.

Cross-task transfer is not automatically beneficial. Joint gains depend on
aligned histories, compatible popularity distributions, and related item
co-occurrence structure; semantic identifiers optimized for one task can
degrade the other \cite{bridging,shi2025gensar,penha2025semanticids}. Our
system therefore shares the intent-to-serving contract---a controlled
vocabulary and typed executable plan---while retaining mode-specific retrieval,
ranking, thresholds, and latency budgets. Whether a plan originating in one
mode causally improves another remains a separate empirical question.

\subsection{Interactive and Steerable Recommendation}

Interactive recommendation predates LLMs. Critiquing systems let users refine
attributes, and conversational systems interleave preference elicitation with
recommendation \cite{chen2012critiquing,jannach2021conversational}. Scrutable
profiles expose preferences for inspection and editing
\cite{balog2019scrutable,mysore2023editable,ramos2024nlprofiles},
while instruction-following models provide language-based control over
recommendations \cite{zhang2025instruction,lu2024controllable}. CTRL-Rec,
RecBot, and SteerEval more directly study natural-language steering
\cite{carroll2025ctrlrec,tang2025recbot,zhou2026steereval}.

Dear Algo's distinction is operational: its intent layer grounds positive,
negative, and compound instructions to production feature identifiers and
compiles them into an inspectable serving plan. Because that plan can outlive
the initiating request,
its lifecycle must handle preference drift and responsiveness to correction
\cite{koren2009temporal,shen2026mars,wang2023negativefeedback}. Provenance and
deletion are additional state-management requirements.

\subsection{Grounded Agentic Retrieval}

The intent layer is related to semantic parsing, entity linking, and
constrained decoding, which separate schema or identifier validity from
semantic correctness
\cite{wang2020ratsql,wu2020entitylinking,scholak2021picard,decao2021genre}.
Query-expansion methods instead generate free-form retrieval text
\cite{gao2023hyde,wang2023query2doc}, while generative retrieval produces item
identifiers directly \cite{tay2022dsi,rajput2023tiger}. The intent layer keeps
the catalog external: an LLM proposes intent, grounding resolves supported
values, and a typed DSL expresses predicates and retrieval parameters.

Agentic recommenders commonly place an LLM control plane over conventional
retrieval and ranking tools
\cite{wang2024recmind,zhao2024toolrec,huang2025interecagent}. Dear Algo follows
this pattern rather than replacing the serving data plane; structured predicates
execute over a neural candidate service, analogous to filtered vector search
\cite{gollapudi2023filtereddiskann,patel2024acorn}. Its bounded
plan-and-execute path is narrower than general tool-using and adaptive
reasoning-and-retrieval loops
\cite{react,trivedi2023ircot,asai2024selfrag}. Classical RAG conditions a
generated answer on passages \cite{rag}; the grounding stage instead retrieves
vocabulary entries before item retrieval. Conversational recommendation and
LLM reranking operate at still different boundaries \cite{wang2022unicrs,rankgpt}.

\subsection{Evaluation Across Paradigms}

Information retrieval (IR) test collections estimate ranking effectiveness
for fixed topics and a fixed corpus; incomplete pools and assessor disagreement
bound that estimand
\cite{buckley2004incomplete,voorhees2000variations}. Offline recommender
benchmarks instead predict held-out behavior under a historical exposure and
candidate protocol, where unexposed items are not valid negatives and sampled
metrics may change conclusions
\cite{shani2011evaluating,krichene2020sampled,
schnabel2016recommendations,joachims2017unbiased}. Neither design by itself
estimates the value of deploying a new cross-mode policy.

Counterfactual estimators require logged propensities and support for the
target actions \cite{li2011unbiased,dudik2011doubly}; those requirements are
not met by the observational study reported here. Randomization most directly
identifies a deployable policy effect \cite{kohavi2009controlled}, but sessions
or clicks still need not measure intent fulfillment, and short experiments can
miss recommendation feedback loops and long-run welfare
\cite{chaney2018algorithmic,mladenov2020optimizing}. We therefore keep request
relevance, admitted-set quality, user behavior, and report actions as
separate estimands.

\subsection{LLM-as-a-Judge for Relevance Evaluation}

LLM judges are measurement instruments, not task-independent oracles.
Pointwise labels, pairwise preferences, and scalar scores answer different
questions \cite{liu2023geval,mtbench,kim2024prometheus}; their behavior varies
with the model, prompt, rubric, scale, decoding, threshold, and population
\cite{faggioli2023perspectives,thomas2024searcher,
arabzadeh2025promptsensitivity,arabzadeh2025benchmarking}. Documented
presentation-order, length, and self-preference effects further motivate
in-domain validation rather than wholesale human replacement
\cite{wang2024fairevaluators,dubois2024length,
panickssery2024selfpreference,soboroff2025dontuse}.

Validation therefore attaches to the complete instrument
\(J=(\text{model/version},\text{prompt},\text{rubric},\text{scale},
\text{decoding},\text{threshold})\). Agreement or system-level correlation
does not calibrate a numeric score or a new operating point
\cite{huang2025empirical,bavaresco2025llms,froebe2025assessors,
dietz2025principles}. For a selective gate, admission precision, coverage, and
relevant recall are distinct; prospective guarantees additionally require a
frozen threshold-selection procedure \cite{jung2025trust}.

Accordingly, our blinded audit validates the production LLM judge scorer at the
strict categorical operating point on the sampled public population. The
paired Top-20 study uses the same scorer but reports fixed-denominator
\emph{judge-qualified yield}, not human precision. The candidate-randomized
study also uses this scorer, but estimates quality among judged admissions
within the serving window rather than quality over an unobserved pre-reranking
slate.

\begin{figure*}[t]
  \centering
\includegraphics[width=0.94\textwidth]{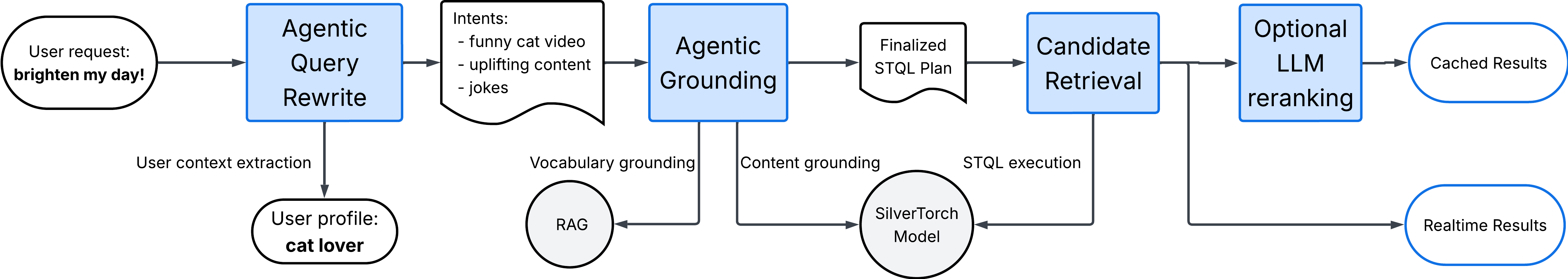}
  \caption{Dear Algo's intent layer compiles heterogeneous intent into a shared
    grounded STQL plan that drives candidate retrieval, scoring, reranking, and
    plan lifecycle operations.}
  \Description{Four intent inputs enter interpretation, controlled-vocabulary
    grounding, and STQL compilation. The resulting plan drives SilverTorch
    candidate retrieval, semantic scoring, optional multimodal reranking, and
    persistence, refresh, expiration, or deletion.}
  \label{fig:architecture}
\end{figure*}

\section{Dear Algo: A Layered Intent Architecture}

Figure~\ref{fig:architecture} summarizes the system's compilation and execution
path.

\subsection{Levels of Unification}

We distinguish three claims:
\begin{enumerate}
  \item \textbf{Infrastructure unification:} modes share orchestration,
    catalogs, or serving components.
  \item \textbf{Representation unification:} modes compile intent into the
    same semantic vocabulary and executable plan.
  \item \textbf{Empirical unification:} information originating in one mode
    causally improves another mode without unacceptable regression.
\end{enumerate}

Dear Algo is designed and implemented to share the first two layers. Catalog
and vocabulary identifiers, intent interpretation, and STQL plans are common;
candidate configuration, ranking, and interaction surfaces remain
mode-specific. The experiments in Sections~\ref{sec:evaluation}
and~\ref{sec:results} do not establish causal empirical unification.

\subsection{Intent Compilation}

Let \(q\) be an optional explicit request, \(h_u\) eligible user context, \(c\)
the immediate interaction context, and \(m\) the discovery mode. The intent
layer receives
\begin{equation}
  x=(q,h_u,c,m), \qquad
  m\in\{\text{explicit},\text{passive},\text{negative},\text{compound}\}.
\end{equation}

It produces a normalized intent \(z\), executable plan \(p\), candidates \(C\),
and ranking \(R\):
\begin{equation}
  \begin{split}
    z &= f_{\mathrm{intent}}(x), \\
    p &= f_{\mathrm{ground}}(z,V), \\
    C &= f_{\mathrm{retrieve}}(p), \\
    R &= f_{\mathrm{rank}}(C,z,h_u),
  \end{split}
\end{equation}
where \(V\) is the controlled semantic vocabulary.

The interpreter emits schema-constrained semantic features rather than an
unconstrained retrieval string. Each feature has an explicit categorized semantic dimension
(e.g., topic, language, location, or freshness), polarity where applicable,
candidate values, and type-specific valid operators. Grounding retrieves
supported vocabulary values and identifiers. The planner then composes
inclusion, exclusion, conjunction, disjunction, language, location, and
freshness constraints into STQL.

\paragraph{Running example.}
Consider the synthetic request \emph{Dear Algo, show me more women's basketball
analysis, but fewer score spoilers}. The interpreter emits two positive
features and one negative feature. Grounding maps the three phrases to
controlled identifiers \(v_1\), \(v_2\), and \(v_3\). A human-readable view of
the resulting plan is
\begin{equation}
  p_{\mathrm{ex}} =
  \operatorname{AND}\!\left(
    \operatorname{TAG}(v_1),
    \operatorname{TAG}(v_2),
    \operatorname{NOT}(\operatorname{TAG}(v_3))
  \right).
\end{equation}
Production serialization replaces readable labels with controlled identifiers
and encodes the Boolean tree for model-query execution. Execution accepts only
typed plans whose operators and grounded values satisfy the serving contract;
unsupported plans abstain or follow the existing fallback path.

\subsection{Execution and Persistence}

The STQL plan is sent to an existing high-throughput candidate generation
service.
Candidates may be scored against the normalized intent and optionally reranked
by a text or multimodal LLM. Expensive stages are restricted to a bounded
slate and can be skipped on passive or latency-sensitive paths. The plan can be
persisted and refreshed against new content without rerunning every reasoning
stage. Architecturally, this separates intent compilation as a control-plane
concern from candidate retrieval as a latency-sensitive data-plane concern.
The design follows a compile-once, execute-many principle: expensive intent
reasoning is amortized across refreshes, while execution remains late-bound to
current inventory. Optional bounded reranking permits latency-sensitive paths
to skip expensive stages without changing the shared plan contract.

\subsection{Implementation Substrate}

Dear Algo's intent-layer implementation is called SilverTorch Agentic
Recommender (STAR). It runs on SilverTorch, a model-based GPU recommendation
serving system that integrates filtering, approximate nearest-neighbor
retrieval, and scoring as model components~\cite{silvertorch}. An explicit
request can be executed immediately and optionally persisted; for feed recommendation, the
persisted or inferred plan can be refreshed against new inventory to produce
intent-conditioned candidates for the existing ranking pipeline. Negative
intent becomes suppression clauses, while compound intent is decomposed into
unit plans whose results are deduplicated and aggregated. Dear Algo therefore
connects search-like requests with recommendation-like continuous retrieval
without replacing the downstream recommender or requiring both modes to share
one ranking model. The explicit surface is not general web search.

\section{Evaluation}
\label{sec:evaluation}

\subsection{Research Questions and Evidence Tiers}

We ask:
\begin{itemize}
  \item \textbf{RQ1:} Can the evaluator support a high-precision categorical
    gate for request-item relevance?
  \item \textbf{RQ2:} Holding the LLM-derived query fixed, does the full
    grounded-tag policy produce more strictly judge-qualified candidate yield
    than the retrieved-tag policy?
  \item \textbf{RQ3:} Within the delivery-eligible window, does randomized LLM
    reranking improve precision-first quality
    among judged admitted items?
  \item \textbf{RQ4:} Is explicit-request initiation followed by higher later
    Threads app usage, and how does the report rate change?
\end{itemize}

The studies answer different questions. Human calibration validates one
measurement operating point. The offline study evaluates paired candidates
from a frozen request-time pool. The candidate-randomized study estimates
judge-labeled quality among admitted impressions within serving common
support. The behavioral study measures within-user change around an explicit
request but lacks a treatment control. We do not combine them into a single
significance claim.

\subsection{Blinded Human Calibration}

We formed a frozen sampling frame of 48,876 non-employee public-mode
request-item evaluation records logged from July 17 through July 24, 2026,
produced by the locked categorical judge configuration and excluding records
used in prior pilot samples. The frame contained 2,212 judge-Irrelevant,
14,138 judge-Partial, and 32,526 judge-Relevant records. We deliberately
sampled 90, 120, and 90 pairs from these strata, respectively. This design
oversamples the rarer Irrelevant and boundary Partial strata for audit
precision; cross-stratum population estimates therefore use inverse selection
weights. Six blinded annotation workbooks yielded 900
initial ratings. Of 300 pairs, 118 had unanimous labels. A seventh independent
blinded rating supplied the final label for 178 evidence-ready disagreements.
Four pairs with insufficient evidence (e.g. due to privacy constraints) were excluded and reported, leaving 296 pairs
from 296 users.

The audited instrument is the locked, pointwise production LLM judge scorer. For
the alignment audit, its logged outputs are represented by the frozen
\texttt{Irrelevant}, \texttt{Partial}, and \texttt{Relevant} strata. The strict
gate admits only \texttt{Relevant}; its primary estimand is exact-Relevant
admission precision:
\begin{equation}
  P(\text{human Relevant}\mid\text{judge Relevant}).
\end{equation}
The production prompt, model version, decoding configuration, and service
details are proprietary and are not disclosed.
At-least-Partial precision is a severe-false-admission sensitivity. Exact
Relevant recall is reported as a diagnostic, not as the product objective.
Strict-gate precision conditions entirely on the judge-Relevant stratum, so
its inverse-probability weight cancels. Metrics that combine judge strata use
the frozen inverse selection probabilities to recover the eligible target
frame. Intervals use 10,000 sampling-stratum-preserving user-cluster bootstrap
replicates with those weights. The zero severe-false-admission count also
receives an exact binomial interval. This primary analysis uses the frozen
adjudicated labels and applies no latent-label or rater-bias adjustment.

Separately, we fit a post-hoc smoothed Dawid--Skene model by
expectation-maximization to examine rater-severity sensitivity. The model uses
the balanced assignments across the six original blinded raters to estimate
per-item latent-class probabilities and treats insufficient evidence as an
abstention response. The selectively assigned seventh rater is excluded. Each
of 10,000 sampling-stratum-preserving bootstrap replicates resamples pairs and
refits the model. This exploratory, model-based sensitivity does not alter the
frozen labels or primary estimate and does not provide a prospectively
calibrated pass threshold.

\subsection{Paired Offline Candidate Study}

The offline dataset contains 100 request events from 98 users. Ninety-two
produced valid continuous judge responses and were aggregated into 72
normalized request-text clusters. Eight malformed JSON responses were excluded
symmetrically from the primary analysis; a sensitivity assigns zero yield to
every arm for those events. Every arm executes on the same SilverTorch
candidate-serving substrate, uses the same request-time available item pool,
and has a fixed Top-20 budget. Infrastructure, inventory, and the slot budget
are held fixed; each configuration may retrieve a different slate.
Table~\ref{tab:arms} defines the displayed configurations.

\begin{table}[t]
  \caption{Paired offline retrieval configuration arms.}
  \label{tab:arms}
  \small
  \begin{tabularx}{\columnwidth}{@{}cYY@{}}
    \toprule
    Arm & Query text & Grounded-tag policy \\
    \midrule
    A & Raw request & Use retrieved canonical tags \\
    B & LLM-derived query & Use retrieved canonical tags \\
    C & LLM-derived query & Full LLM-selected grounded-tag policy \\
    \bottomrule
  \end{tabularx}
  \begin{minipage}{\columnwidth}
    \footnotesize\textit{Note:} All displayed arms use the same substrate and
    request-time inventory.
  \end{minipage}
\end{table}

All displayed configurations use tag grounding and compile the result through
STQL. C--B is therefore a configuration-level contrast between two
grounded-tag policies after holding the LLM-derived query fixed; it does not
isolate any individual grounding mechanism.

For request \(i\) and arm \(a\), the fixed-denominator outcome is
\begin{equation}
  Y_{i,a}@20 =
  \frac{1}{20}\sum_{j=1}^{20}\mathbf{1}[s_{i,a,j}\geq 0.8].
\end{equation}
Let \(Q_{i,a}\) denote the numerator above. Table~\ref{tab:offline-results}
reports the primary outcome as both Yield@20, \(Y_{i,a}@20\), and qualified
candidates@20, \(Q_{i,a}\); these are the same fixed-denominator outcome in
rate and count units, respectively.
Here \(s_{i,a,j}\) is the continuous judge score on \([0,1]\). The value 0.8
was a pre-existing internal product-review quality bar used in the initial
offline evaluation, before C--B was selected as the focal paper contrast. We
retained it post hoc as the qualification gate for the precision-first
analysis. The cutoff was not tuned to the C--B contrast, selected from human
labels, or used as a production admission threshold. We report sensitivity at
the stricter thresholds 0.9 and 1.0.
Unjudged candidates, missing summaries, and short-slate positions receive zero
yield. Each metric is first computed per request event; duplicate events are
then averaged within normalized request-text clusters, and clusters receive
equal weight. We report 20,000 cluster-bootstrap replicates. The score comes
from the same production LLM judge scorer used in the human alignment. Because the
outcome counts score-qualified slots rather than human-labeled relevance among
admissions, we call it \emph{judge-qualified yield}, not human precision.

\subsection{TTL-Aligned Candidate-Randomized Reranker Study}

We analyze public-mode admitted impressions logged from July 17 through July
24, 2026. Candidates within a request were randomized between the standard
candidate source (reranker off) and the direct-response LLM-reranker source
(reranker on). The request is therefore the analysis block. Content is
deduplicated within each request and arm before computing outcomes.

Delivery-stack validation established that the reranker-on source has a
three-day time-to-live (TTL), whereas the off source can continue dynamic
retrieval through the seven-day Dear Algo lifetime. The primary comparison is
therefore restricted to the common-support window
\(0 \leq \text{request-to-impression age} < 72\) hours. This restriction was
adopted after operational validation and was not selected from the quality
outcomes. The observable age begins at Dear Algo request authoring, while the
serving TTL begins at message receipt, so the 72-hour window is a conservative
proxy for the exact eligibility clock.

The window contains 93,972 logged rows across 4,321 requests. Of 1,531
paired-arm requests from 1,417 users, 1,346 requests from 1,278 users have at
least one categorical LLM-as-a-judge label in both arms. After request-arm-item
deduplication, the paired cohort contains 50,922 off candidates (31,099 judged)
and 3,770 on candidates (2,928 judged); 845 requests from 823 users are fully
judged in both arms.

The two sources retained their production candidate budgets. Because the
agentic reranker-on path has higher serving computational cost and latency, it generated a
substantially smaller candidate set per request; the 50,922-versus-3,770
imbalance is therefore expected by design rather than an allocation target or
attrition. Candidate randomization occurred within request, but the paths did
not operate over an equal-sized frozen slate. The study therefore compares
end-to-end serving paths under their production candidate budgets.

For each request and arm, Exact-Relevant share is the fraction of judged
admitted candidates labeled Relevant, and false-admission rate is the fraction
labeled Irrelevant. Partial is neither a strict success nor a false admission.
We first average candidates within request, then requests within user, and
report the mean paired user contrast. Intervals use 20,000 user-cluster
percentile-bootstrap replicates. Fully judged requests and same-request-session
comparisons are sensitivities.

We also compute arbitrary-MNAR bounds over all 1,531 paired requests. For the
lower Exact-Relevant bound, every unjudged on candidate is treated as not
Relevant and every unjudged off candidate as Relevant; the upper bound reverses
those assignments. False-admission bounds analogously assign every missing
label adversarially. Each endpoint is reaggregated request-to-user and
bootstrapped. Because rejected candidates and a frozen pre-reranking slate are
unavailable, the study estimates admitted-set quality, not recall, NDCG, or
rank movement.

\subsection{Observational Explicit-to-Feed Study}

We identify each user's first eligible explicit-request day \(D\) after at
least 28 days without another eligible request. We compare home-feed behavior
in \(D-7,\ldots,D-1\) with \(D+1,\ldots,D+7\). The index day is excluded
because the request timestamp is not trusted. We analyze three entry-point
cohorts separately. In \emph{public creation}, a user authors a public Threads
post containing a Dear Algo request that other people can see. In
\emph{public repost}, a user reposts another person's public Dear Algo request,
adopting that request for the user's own feed. In \emph{private persisted
intent}, a user adds a Dear Algo request privately, visible only to that user.
These entry points differ in social visibility and request origination and may
therefore select users with different motivations and baseline behavior.
Separate reporting prevents cohort heterogeneity from being hidden by a pooled
average, but it does not remove within-cohort self-selection.

The log has no native session identifier. We define a proxy session as a
user/day/hour group and order its events by logged event time. The topline is
the paired per-user change in proxy-session count. The negative-feedback
endpoint is the number of report actions per 1,000 exposures, computed over
the first 20 event-ordered exposures in each proxy session. Standard errors are
clustered by index date.

All endpoints use two-sided 95\% index-date-clustered intervals. Public
creation and repost each have 54 index-date clusters; private persisted intent
has 52. The intervals are descriptive and are not adjusted across request
types or guardrails; no formal non-inferiority decision is assigned. A
report-rate interval that includes zero is inconclusive, not evidence of
non-inferiority: such a claim would require a prespecified maximum acceptable
increase \(\delta\) and an upper confidence bound below that margin.

\section{Results}
\label{sec:results}

\subsection{Human Calibration Supports a Narrow Strict Gate}

Table~\ref{tab:human-calibration} reports the strict-gate estimates.

\begin{table}[t]
  \caption{Human calibration of the strict categorical relevance gate
    (\(N=296\) evaluable audit pairs; \(N_{\mathrm{admitted}}=89\)).}
  \label{tab:human-calibration}
  \small
  \begin{tabularx}{\columnwidth}{@{}Yrr@{}}
    \toprule
    Gate metric & Estimate & 95\% interval \\
    \midrule
    Exact precision\textsuperscript{a}
      & \textbf{0.9438} & [0.8876, 0.9888] \\
    At-least-Partial precision\textsuperscript{b}
      & 1.0000 & [0.9594, 1.0000] \\
    \bottomrule
  \end{tabularx}
  \begin{minipage}{\columnwidth}
    \footnotesize
    \textit{Note:} The primary metric is shown in bold.\\
    \textsuperscript{a}\,The strict gate admits judge-Relevant items and counts
    human-Relevant items as successes: 84/89. Bootstrap interval.\\
    \textsuperscript{b}\,The same 89 admissions are successful when human-Partial
    is also acceptable: 89/89. Exact Clopper--Pearson interval.
  \end{minipage}
\end{table}

Exact-Relevant precision was 84/89 = 0.9438 (95\% CI [0.8876, 0.9888]). The
remaining five admissions were human-Partial and none was human-Irrelevant,
yielding an at-least-Partial precision of 89/89 = 1.0000 (exact 95\% CI
[0.9594, 1.0000]). In the stratified audit sample, the strict gate admitted 84
of 176 human-Relevant pairs; it assigned 76 to Partial and 16 to Irrelevant.
This operating point trades coverage for admission precision, aligning with
conservative persistent-feed admission rather than exhaustive retrieval.

Across the eligible frame, the locked judge labeled 32,526 of 48,876 records
Relevant, corresponding to a 66.55\% gate admission rate. Because the audit was
stratified by judge label, its raw 89/296 admission fraction does not estimate
this rate.

The post-hoc six-rater model-based sensitivity estimated exact precision at 0.8799
[0.7886, 0.9515], severe false admission at 0.0208 [0.0049, 0.1217], and
at-least-Partial precision at 0.9792 [0.8783, 0.9951]. The seventh rater agreed
with all 21 original majorities among 34 contested admissions and resolved 13
ties as nine Relevant and four Partial. An item-equal valid-vote check estimated
exact precision at 0.8561 [0.7989, 0.9080] and Irrelevant share at 0.0379
[0.0114, 0.0712]. These sensitivities are reported beside, not instead of, the
frozen labels.

The result is descriptive, not a preregistered pass. The precision-first
estimand and threshold were clarified after label inspection. Initial human
reliability was also weak: pairwise exact agreement was 0.677 and quadratic
weighted kappa was 0.250 before the final-label rule.

\subsection{The Full Configuration Yields More Judge-Qualified Candidates}

Table~\ref{tab:offline-results} reports the paired configuration comparison.

\begin{table*}[t]
  \caption{Judge-qualified yield across score thresholds over \(N=72\)
    normalized request-text clusters with a fixed \(K=20\) slot budget.}
  \label{tab:offline-results}
  \small
  \setlength{\tabcolsep}{3pt}
  \begin{tabularx}{\textwidth}{@{}YrrrZZ@{}}
    \toprule
    Metric / score threshold & A: raw & B: derived & C: full
      & C--B [95\% CI] & C--A [95\% CI] \\
    \midrule
    Yield@20, score \(\geq 0.8\)
      & 0.279 & 0.331 & 0.386
      & \textbf{+0.056 [0.006, 0.106]}
      & \textbf{+0.107 [0.030, 0.186]} \\
    Yield@20, score \(\geq 0.9\)
      & 0.240 & 0.281 & 0.337
      & \textbf{+0.056 [0.014, 0.099]}
      & \textbf{+0.097 [0.021, 0.175]} \\
    Yield@20, score \(\geq 1.0\)
      & 0.170 & 0.200 & 0.249
      & \textbf{+0.049 [0.011, 0.088]}
      & \textbf{+0.079 [0.015, 0.146]} \\
    \midrule
    Qualified candidates@20, score \(\geq 0.8\)
      & 5.58 & 6.61 & 7.73
      & \textbf{+1.11 [0.12, 2.12]}
      & \textbf{+2.15 [0.59, 3.73]} \\
    \bottomrule
  \end{tabularx}
  \begin{minipage}{\textwidth}
    \footnotesize\textit{Note:} The main reported score threshold is 0.8; 0.9
    and 1.0 are stricter-threshold sensitivities. Yield@20 divides the number
    of qualified candidates by the fixed 20-slot denominator; the final row
    gives the equivalent count units at 0.8. C--B is the focal comparison and
    C--A is contextual. Boldface marks contrasts whose unadjusted 95\% interval
    excludes zero. All intervals are cluster-bootstrap; all rows use 72
    clusters.
  \end{minipage}
\end{table*}

At the 0.8 threshold, C reached 7.73 qualified candidates@20, adding 1.11
[0.12, 2.12] relative to B in the focal comparison and 2.15 [0.59, 3.73]
relative to A in the contextual comparison. Across all three thresholds in
Table~\ref{tab:offline-results}, both the C--B and C--A intervals excluded zero.
The C--B Yield@20 gains were +0.056, +0.056, and +0.049 at thresholds 0.8,
0.9, and 1.0, respectively; the corresponding C--A gains were +0.107, +0.097,
and +0.079. Assigning zero yield to every arm for all eight malformed responses
gives +0.0484 [0.0004, 0.0965]. The supported result is a configuration-level
increase in fixed-denominator judge-qualified yield, not higher human precision
or an independent grounding effect.

\subsection{Lower False Admission Among TTL-Aligned Judged Admissions}

Table~\ref{tab:reranker-results} reports judged-admission quality within the
common-support window.

\begin{table*}[t]
  \caption{Candidate-randomized reranker quality within the delivery-eligible
    first 72 hours. Outcomes are evaluator-defined shares among judged
    admitted candidates.}
  \label{tab:reranker-results}
  \small
  \begin{tabularx}{\textwidth}{@{}lYcrrZ@{}}
    \toprule
    Analysis & Outcome & Users / blocks & Reranker Off & Reranker On
      & Difference in points [95\% CI] \\
    \midrule
    Primary paired request
      & Judge-Irrelevant share & 1,278 / 1,346
      & 4.78\% & 2.80\% & \textbf{-1.97 [-3.02, -0.94]} \\
    Primary paired request
      & Exact-Relevant share & 1,278 / 1,346
      & 65.58\% & 67.82\% & \textbf{+2.24 [0.08, 4.41]} \\
    Fully judged request
      & Judge-Irrelevant share & 823 / 845
      & 5.65\% & 3.19\% & \textbf{-2.45 [-3.92, -0.99]} \\
    Fully judged request
      & Exact-Relevant share & 823 / 845
      & 63.46\% & 66.52\% & \textbf{+3.06 [0.19, 5.86]} \\
    Same request-session
      & Judge-Irrelevant share & 949 / 1,306
      & 3.12\% & 2.85\% & -0.27 [-1.45, 0.91] \\
    Same request-session
      & Exact-Relevant share & 949 / 1,306
      & 66.16\% & 68.86\% & +2.69 [-0.12, 5.53] \\
    \bottomrule
  \end{tabularx}
  \begin{minipage}{\textwidth}
    \footnotesize\textit{Note:} Primary pairs requests with at least one judged
    admission in each arm. Fully judged restricts to requests whose admitted
    candidates are all judged in both arms. Same request-session pairs the
    arms within a common request-session block. Users / blocks reports users
    and requests for the first two analyses, and users and request-session
    blocks for the last. Boldface marks contrasts whose 95\% interval excludes
    zero. Intervals use 20,000 user-cluster bootstrap replicates. Lower
    Judge-Irrelevant share is better.
  \end{minipage}
\end{table*}

Among judged admissions, the primary reranker-on minus off differences were
-1.97 points [-3.02, -0.94] in Judge-Irrelevant share and +2.24 points
[0.08, 4.41] in Exact-Relevant share. The fully judged sensitivity showed
corresponding differences of -2.45 points [-3.92, -0.99] and +3.06 points
[0.19, 5.86]. In the same request-session sensitivity, the differences were
-0.27 points [-1.45, 0.91] and +2.69 points [-0.12, 5.53], respectively, with
both intervals including zero. Within the shared 72-hour window, the
reranker-on path improved both primary quality outcomes among judged admissions
under production budgets.

\subsection{Higher Later Usage with No Significant Change in Reports}

Table~\ref{tab:behavioral-results} reports the primary all-eligible cohorts.
Report rates are scaled per 1,000 eligible exposures. All intervals are
two-sided and clustered by index date.

\begin{table*}[t]
  \caption{Seven-day observational usage and report-rate changes after
    explicit-request initiation.}
  \label{tab:behavioral-results}
  \small
  \begin{tabularx}{\textwidth}{@{}YrrrZrrrZ@{}}
    \toprule
      & & & & & \multicolumn{4}{c}{Reports per 1,000 exposures} \\
    \cmidrule(lr){6-9}
    Entry point & Users & Pre sess. & Post sess.
      & Relative usage lift [95\% CI] & Pre & Post & \(\Delta\)
      & Two-sided 95\% CI \\
    \midrule
    Public creation
      & 40,376 & 37.26 & 39.63
      & \textbf{+6.36\%} [5.67\%, 7.04\%]
      & 0.036 & 0.027 & -0.009 & [-0.021, 0.003] \\
    Public repost
      & 47,412 & 35.98 & 37.47
      & \textbf{+4.15\%} [3.30\%, 4.99\%]
      & 0.018 & 0.017 & -0.002 & [-0.009, 0.005] \\
    Private persisted intent
      & 43,608 & 22.45 & 24.39
      & \textbf{+8.61\%} [7.74\%, 9.48\%]
      & 0.016 & 0.015 & -0.001 & [-0.007, 0.005] \\
    \bottomrule
  \end{tabularx}
  \begin{minipage}{\textwidth}
    \footnotesize\textit{Note:} Sessions are average seven-day proxy-session
    counts per user around index day \(D\). Reports are measured over the first
    20 event-ordered exposures per session. Intervals are unadjusted two-sided
    95\% index-date-clustered intervals.
  \end{minipage}
\end{table*}

All three cohorts show higher post-period usage. Report-rate point estimates
are lower in every cohort, but every two-sided interval includes zero.
Excluding users with a repeat request retains the session pattern, while every
report-rate interval again includes zero.

Together, these results show that explicit-request initiation is followed by
higher Threads app usage, while report-rate changes are not statistically
significant.

\section{Discussion}

\subsection{Unification Is Layered, Not Binary}

Dear Algo implements shared vocabulary, plan, and candidate interfaces for
explicit and recommendation-derived intent. This architecture also illustrates
why the two modes should not automatically share every operating point. An
explicit request may tolerate a broad candidate stage followed by ranking; persistent
recommendation intent may require a more conservative admission gate. A
modular shared layer can limit negative transfer while still enabling
cross-mode state.

\subsection{Bridging IR and RecSys Evaluation}

The offline candidate study resembles pooled IR evaluation: the request-time
inventory and 20-slot budget are held fixed, while each configuration may
retrieve a different slate that is compared through paired resampling. The
candidate-randomized study estimates a serving-path
effect, but only on admitted, judge-labeled impressions inside arm eligibility;
it cannot recover recall over the pre-reranking universe. The behavioral study
resembles recommender-system analysis: it operates over users and time and
measures sessions and report actions. These evaluations answer different
questions. A unified system needs a contract that records request relevance,
user outcomes, latency, computational cost, and safety rather than selecting whichever metric
is favorable.

\subsection{Future Work: Empirical Unification}

The next step is to test whether carrying an explicit request into feed
recommendation causally improves intent fulfillment. Eligible users or
requests should be randomized between stored-plan enabled and disabled
conditions while holding inventory, retrieval budgets, and downstream ranking
fixed. The primary endpoint should be human intent fulfillment or a separately
validated categorical evaluator on pooled, blinded candidates; app usage and
report rate are secondary outcomes. Assignment, plan, inventory, fallback,
and candidate-slate versions should be logged to support reproducible
analysis.

\section{Limitations and Responsible Use}

The study has several principal limitations. First, the explicit-to-feed
behavioral analysis is observational and uses hour-level proxy sessions; user
self-selection, temporal trends, regression to the mean, and changes in feed
composition may contribute to the observed differences. Second, all offline
configurations in the paired candidate study use grounded tags and STQL, so
that study cannot isolate their individual contributions. Third, human
calibration covers public request-item pairs and
does not establish private, per-intent, or multimodal validity; its post-hoc
latent-label sensitivity is also prior- and rater-dependent. Fourth, the
precision-first endpoints were clarified after initial result inspection, and
the behavioral intervals are not multiplicity-adjusted. Fifth, the reranker study
observes only admitted candidates: its TTL-aligned request-age window is a
proxy for serving eligibility, judgment coverage and candidate composition
differ by arm, arbitrary-MNAR bounds span both directions, and recall or
pre-reranking rank movement cannot be estimated.

Persisted explicit or inferred intent may reveal sensitive interests or amplify
an unwanted feedback loop. Production-system confidentiality and user-privacy
constraints prevent release of implementation details and user-level logs; all
reported results are aggregate, and we do not reproduce user requests. A
deployable system requires
data minimization, access control, expiration, deletion, and user correction.
Negative preferences require collateral-suppression audits. Language, locale,
visual-content, and low-history slices also require separate evaluation.

\section{Conclusion}

Dear Algo uses a shared, grounded, and executable intent layer for explicit and
recommendation-like discovery while preserving task-specific serving choices.
Human calibration supports a conservative categorical relevance gate, and an
offline paired study finds higher judge-qualified candidate yield for the full
retrieval configuration. Within its TTL-aligned delivery window, a
candidate-randomized serving-path study finds a 1.97-percentage-point lower
user-weighted judge-Irrelevant share and a 2.24-percentage-point higher
Exact-Relevant share among judged admissions for the reranker-on path under
production budgets. Across three observational cohorts, explicit-request
initiation was followed by 4.15\%--8.61\% higher subsequent Threads app usage
as measured by proxy-session counts, while all report-rate confidence intervals
included zero. Future randomized plan-on/off studies will estimate effects on
intent fulfillment, with app usage and report rate as secondary outcomes.

The results support representation-level unification and precision-first
evaluation, not yet causal cross-task transfer.

\begin{acks}
We thank Sakariya Ahmed, Nick Joodi, Shuting Wang, Ivan Ji, Dan Day, Nadav
Lavon, Laura Javier, Francis Luu, Sophie Theis, Justin Wang, Hannes Verlinde,
Jimmy Saade, Ron Edelstein, Bo Chan, Bernard Nuesa, Kerri MacDonald, Chris
Connolly, Peter Cottle, Keke Zhai, Danyang Wang, Liang Wang, Yijie Deng, Zhen
Wang, Eric Kim, Yiming Ma, Hong Wu, Peng Xia, Hongzhang Yin, Min Ni, Sharon
Zhang, Chae Yoo, Yang Ye, Hong Li, Christopher Schrader, Assaf Cohen, Anuj
Desai, and Rex Cheung for their contributions, feedback, and support.
\end{acks}

\end{document}